\pdfoutput=1
\relax
\documentclass[letterpaper]{article} 
\usepackage{aaai22}  
\usepackage{times}  
\usepackage{helvet}  
\usepackage{courier}  
\usepackage[hyphens]{url}  
\usepackage{graphicx} 
\usepackage{natbib}  
\usepackage{caption} 
\DeclareCaptionStyle{ruled}{labelfont=normalfont,labelsep=colon,strut=off} 
\usepackage{algorithm}
\usepackage{algorithmic}

\usepackage{newfloat}
\usepackage{listings}
\floatstyle{ruled}
\newfloat{listing}{tb}{lst}{}
\floatname{listing}{Listing}
\title{Towards Automated Error Analysis: \\Learning to Characterize Errors}
\author {
    Tong Gao, 
    Shivang Singh,
    Raymond J. Mooney 
}
\affiliations {
    Department of Computer Science\\
    University of Texas at Austin\\
    \{gaotong, ssingh, mooney\}@cs.utexas.edu
}

\begin{document}

\maketitle

\begin{abstract}
Characterizing the patterns of errors that a
system makes helps researchers focus future development on increasing its accuracy and robustness. We propose a novel form of "meta learning'' that automatically learns interpretable rules that characterize the types of errors that a system makes, and demonstrate these rules' ability to help understand and improve two NLP systems. Our approach works by collecting error cases on validation data, extracting meta-features describing these samples, and finally learning rules that characterize errors using these features. We apply our approach to VilBERT, for Visual Question Answering, and RoBERTa, for Common Sense Question Answering. Our system learns interpretable rules that provide insights into systemic errors these systems make on the given tasks. Using these insights, we are also able to ``close the loop'' and modestly improve performance of these systems.
\end{abstract}

\section{Introduction}

Although deep-learning systems have made remarkable progress in recent years, systems still make a significant number of errors on complex and diverse tasks. An important step in engineering highly-accurate, robust
systems is error analysis, which has been defined as the process of 
examining development set examples misclassified by the algorithm and understanding the underlying causes of those missclassifications. This process helps engineers prioritize critical problems and prompts them in the direction of handling the problems \citep{solegaonkar19}. 

Unfortunately, such
analysis usually needs manual inspection and reasoning, which is an onerous, time-consuming and hit-or-miss process. 
Pure manual analysis may lead to a biased conclusion, as common features appearing in both successful and failed classifications could be misunderstood as the root cause of failure \citep{rondeau-hazen-2018-systematic}. While other works have proposed improved approaches to error analysis \citep[e.g.][]{wu-etal-2019-errudite,kahng2017cti, slicing}, some are not scalable to large-scale datasets, hindering their practical application, while others require a deep understanding of the errors even before conducting their proposed method.  In contrast, our pipeline (1) makes weak prior assumptions about the attributes of failures, (2) is scalable to large-scale datasets, (3) groups errors with interpretable and globally unbiased rules, enabling fast manual inspection and analysis. Developers can benefit from such a systematic view and develop patches addressing different problems disclosed by different sets of rules.

\begin{figure*}[t]
    \centering
    \includegraphics[width=0.9\textwidth]{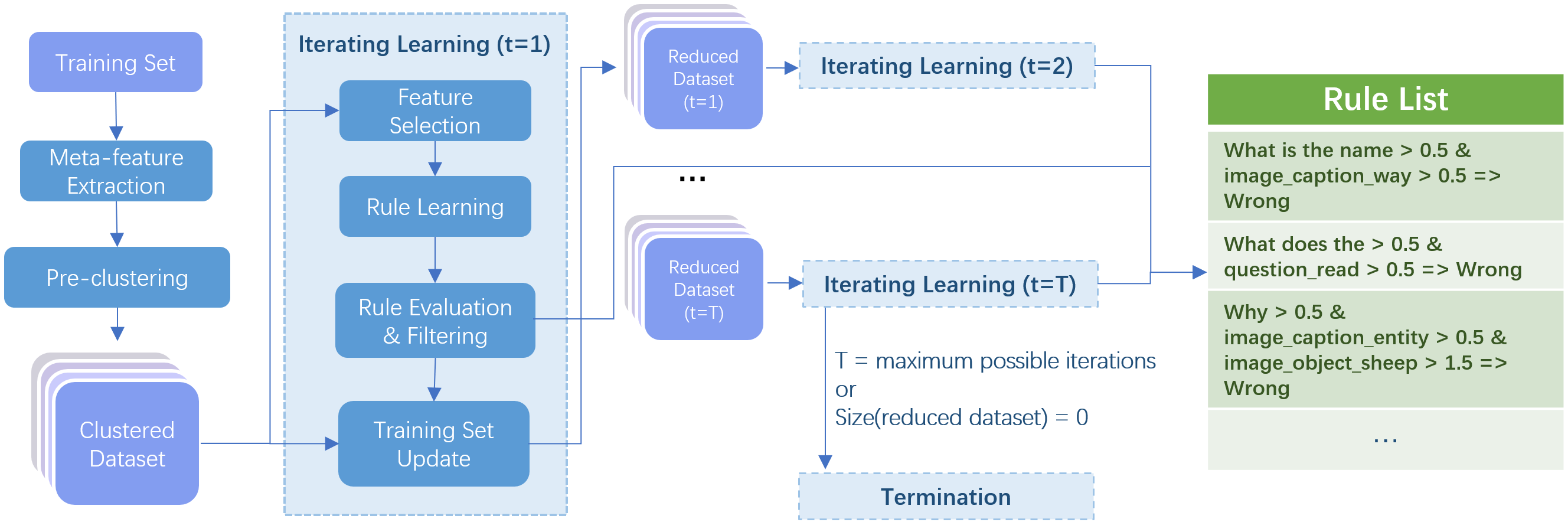}
    \caption{Pipeline Overview}
    \label{fig:pipeline}
\end{figure*}

Specifically, we propose using machine learning to help automate error analysis by inducing interpretable rules that characterize the
errors that a system makes. First, our approach runs a predictive model on a set of held-out validation data and
records which examples the system classifies correctly or incorrectly. Next, we characterize each example using a set of
“meta-features” that describe the problem (e.g. for question answering, a meta-feature could be: question
starts with “How many”). Finally, we learn interpretable rules from this data which predict failure using
these meta-features (Figure \ref{fig:pipeline}).
For example, one rule could be: ``If the question starts with `How many’ and the answer
is greater than 2, then it is probably wrong.'' 
In fact, this is an example of a rule learned by a prototype of our
approach when analyzing the errors made by a Visual Question Answering (VQA) system that does not include a specific mechanism for counting problems.

Developers can then browse the errors grouped by the rules and gain insight from the errors and their description. This can guide them to engineer improvements to the system that increase its accuracy and robustness.
They can also help developers and users build a
“mental model” of the system that enables them to better predict its performance and gauge when and when not
to trust it.

To demonstrate the effectiveness of our approach, we present results applying our approach to ViLBERT \cite{lu2019vilbert} for Visual Question Answering v2.0 (VQA) \cite{Goyal2018MakingTV}, and RoBERTa \cite{liu2019roberta} for CommonsenseQA (CSQA) \cite{talmor-etal-2019-commonsenseqa}. We induce human-interpretable rules providing insights into systemic errors these systems make on these tasks. We then show how to ``close the loop'' and modify the systems to improve their performance using some of these insights.

\section{Automatic Error Characterization}

This section describes the pipeline for our approach to automatic error characterization.  It first runs a pre-trained system on development data not used during
training and collects the predictions that it makes. Next, these examples are described using a set of extracted meta-features that characterize properties of the problems. Finally, we run a rule learner that uses these meta-features to characterize the model's mistakes. Figure \ref{fig:pipeline} provides a visualization of this pipeline.

\subsection{Meta-feature Extraction}

Meta-feature engineering is a domain-specific process requiring understanding of the problem as well as the properties that affect model behavior. 
For our two QA tasks, language tokens in the questions and the gold answers are the most important meta-features. For images in VQA, we added meta-features representing objects in images detected by YOLOv3 \citep{yolov3} as well as generated image captions \cite{luo2018discriminability}, which may encode extra properties and relations for the objects in the image.

\begin{figure}[h]
    \centering
    \includegraphics[width=\columnwidth]{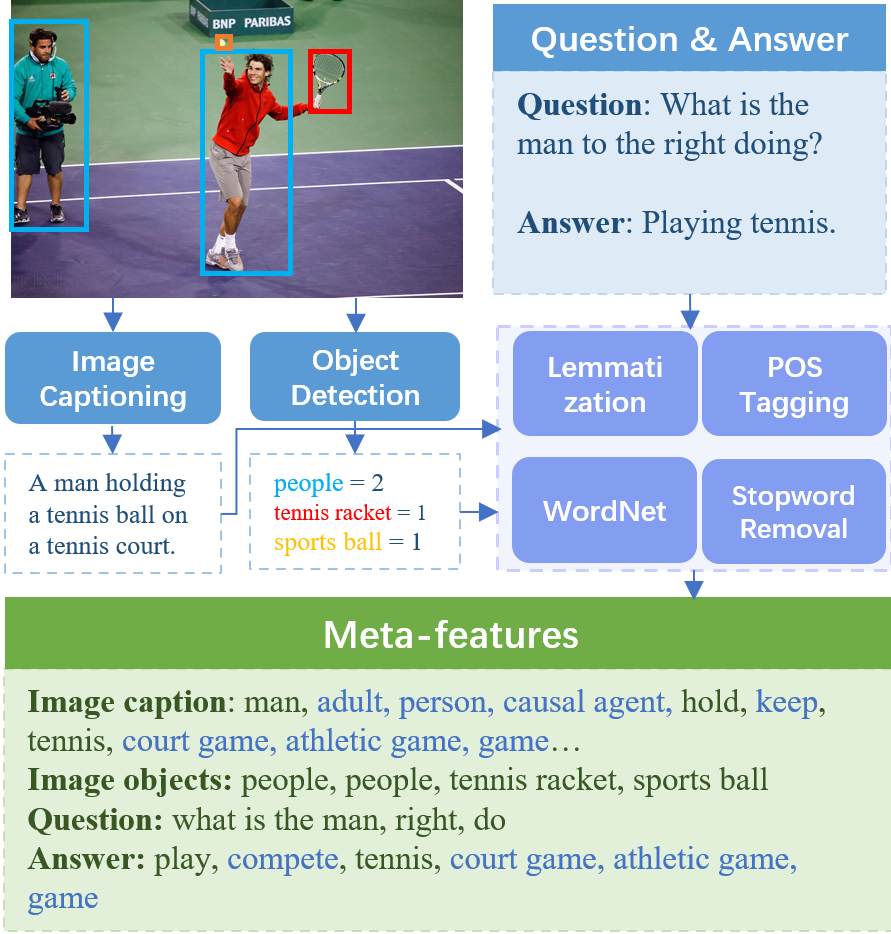}
    \caption{Meta-feature extraction process. Hypernyms in meta-features are marked in blue.}
    \label{fig:meta-feat}
\end{figure}

All the text tokens are then lemmatized and part-of-speech tagged, providing additional meta-features. We lookup all nouns and verbs in WordNet \citep{miller1995wordnet} and add their hypernyms up to 4 levels as additional meta-features. Hypernyms help group similar features across examples and allow more abstract characterizations of error cases. Finally, stopwords are removed and all meta-features are represented by a sparse vector, where each entry represents the occurrence frequency of one meta-feature in an example. Figure \ref{fig:meta-feat} gives an overview of this process for VQA.

\subsection{Pre-clustering}

Characterizing a large dataset is  difficult for existing rule learners due to issues scaling to large numbers of examples and features.
Hence, motivated by discriminative clustering \cite{Bansal2014TowardsTS}, we first cluster the dataset into several smaller sub-datasets and run the rule learner on each cluster separately. This pre-clustering step speeds up the rule extraction process and helps arrange the resulting rules in a semantically meaningful way.

Following the prior work \cite{Bansal2014TowardsTS}, we run k-means on the error cases (i.e. positive examples) and build 2 clusters for CSQA. Both of them are combined with all negative examples, creating 2 smaller overlapping sub-datasets for rule learning. Similarly, we run k-means on VQA error cases and found that words that typically represent question types (e.g. ``why") are all close to cluster centers. To obtain finer-grained clusters that better divide the tremendous amounts of data, we proceeded with this observation and categorize the dataset into 65 clusters by question types which are given in the official annotations for this data.


\subsection{Rule Learning}


Finally, we automatically induce a model that predicts success or failure on an example given its meta-features. For this task we used a rule learner which provides us interpretable insight into the errors made by the target model. Our system utilizes SkopeRules, a Python based machine learning library that ``aims to learn logical and interpretable rules'' \cite{Gardin2017}. While other interpretable classifiers such as simple linear models could also have been used, we present the rationale for choosing rule learning in Section \ref{sec:comp}, where we compare our approach to other learned models trained to predict errors. 

\subsubsection{Feature Selection}


In practice, the raw extracted meta-feature dataset is high dimensional, sparsely populated, and noisy. This prevents the rule learner from generating effective rules, justifying the necessity of dimensionality reduction. Following \citet{yang1997comparative}, we used a chi-square feature selection method as it is an efficient way to reduce the number of features. We performed grid search on the possible number of dimensions and selected 100 meta-features as the best value.

\subsubsection{Iterative Learning}
In order to make the rule learner more efficient and effective, we run it multiple times with a high precision threshold ($60\%$) for the rules it produces, redoing feature selection at each iteration to allow new rule sets to potentially focus on a new set of important features.   
At the beginning of each iteration, chi-square is used to select features and SkopeRules is run on the current training set to generate a set of high-precision rules for identifying error cases. We then remove the cases covered by these rules and repeat this process until no more error cases can be characterized with high precision, or until a maximum number of iterations is reached, which is 50 in our configuration.

\subsubsection{Rule Evaluation and Filtering}
\label{sec:ruleeval}
To ensure learned rules accurately predict errors and do not just overfit the development set, we created a 90/10 training/test split from the original development set for the rule learning process. Rules are then learned from the training subset and evaluated on the test subset. Any rule that is not at least $60\%$ accurate on the test split is removed from the final rule set.

\section{Experimental Results}
\label{sec:exp}

We ran our pipeline on ViLBERT pretrained on VQA, and RoBERTA finetuned on CSQA. We collected 404 and 81 rules covering 14.1\% and 60.8\% of error examples for two tasks respectively.
Examining the rules learned from our approach, we were able to gain deeper insights into both tasks.
Below we highlight and discuss some of the error patterns that we have detected.

\subsection{ViLBERT on VQA}

\subsubsection{OCR Problems}

\begin{figure*}[h]
    \centering
    \includegraphics[width=0.9\textwidth]{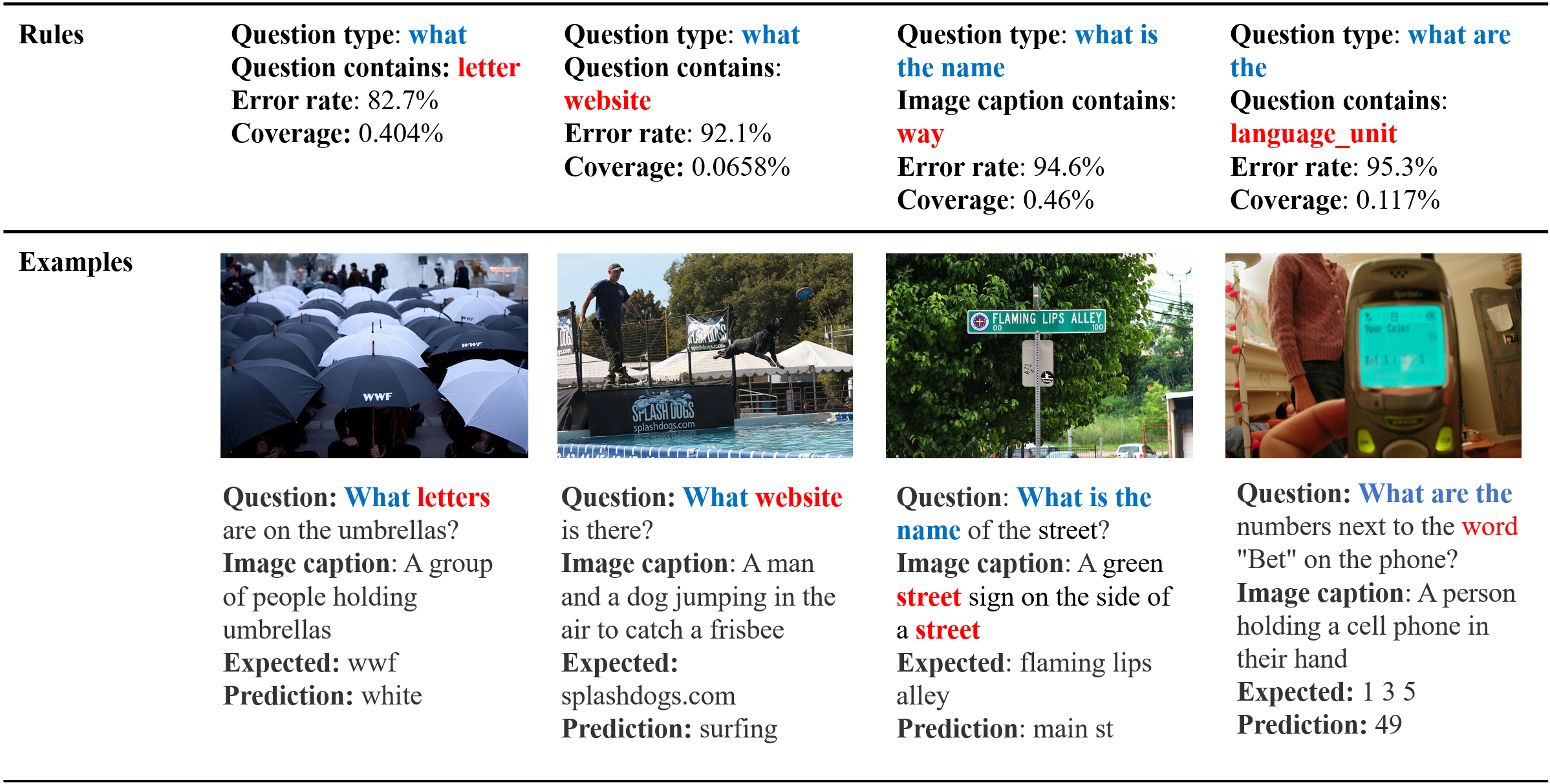}
    \caption{OCR problem examples. Words (or their hypernyms) mentioned by the rule are marked in red. Question types are marked in blue. }
    \label{fig:ocr}
\end{figure*}

The major category of errors that we identified involves reading text from images, therefore requiring Optical Character Recognition  (OCR). Sample rules together with a covered problem instance are shown in Figure \ref{fig:ocr}. For example, when a question, starting with ``what'', asks about a ``letter,'' ViLBERT has an 82.7\% chance of getting it wrong. 
Similarly, questions involving URLs and dates have low accuracy. 
 Sign-reading problems are another typical OCR-type problem in VQA; as shown by the third example involving road signs.  
 Some challenging OCR problems also exist in VQA, as shown by the fourth example, which requires reasoning about texts’ positioning. 
 Since the lack of OCR ability is the most significant problem  uncovered by our system, we propose a mixed architecture to alleviate it in Section\ \ref{sec:loop}.

\subsubsection{Referring Expression Problems}
\begin{figure*}[h]
    \centering
    \includegraphics[width=0.9\textwidth]{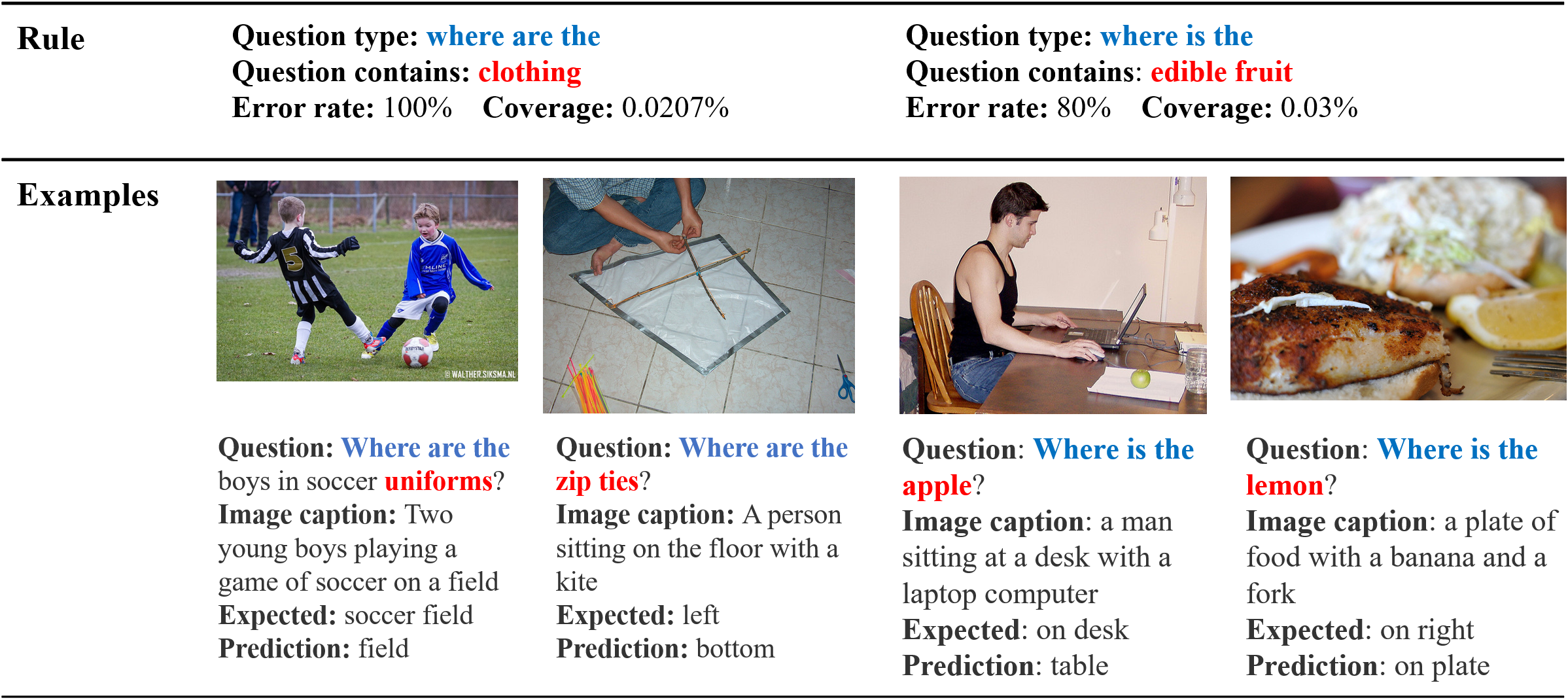}
    \caption{Referring expression problem examples}
    \label{fig:refer}
\end{figure*}
Several rules indicate that examples that require describing the position of an object are likely to cause failure, even if the answer is reasonable. Due to VQA’s open-ended nature, VQA accuracy, the standard VQA evaluation metric shown below, frequently incorrectly scores answers to such questions: 
$$\mbox{VQA Accuarcy} = \min\left(\frac{n}{3}, 1\right)$$
where $n$ is the number of gold answers exactly matched by the generated answer. While there are 10 gold answers as candidates, in practice, many of the 10 gold answers are identical and thus provide low tolerance to reasonable answer variants.

The examples in Figure \ref{fig:refer} illustrate this problem for ``where"-type questions, where the answers are reasonable but counted as wrong.  This illustrates a problem with the scoring method rather than with the model. One proposed solution is Alternative Answer Sets \cite{luo-etal-2021-just}, an improved metric relying on WordNet to recognize synonyms of a gold answer. This might solve the first and third problem in Figure\ \ref{fig:refer}, but it would not solve the other ones. In general, this requires determining whether two {\it referring expressions} denote the same object in an image, a difficult language-vision problem \cite{ref-expr}.

\subsubsection{Time-reading Problems}
\begin{figure}[h]
    \centering
    \includegraphics[width=0.9\columnwidth]{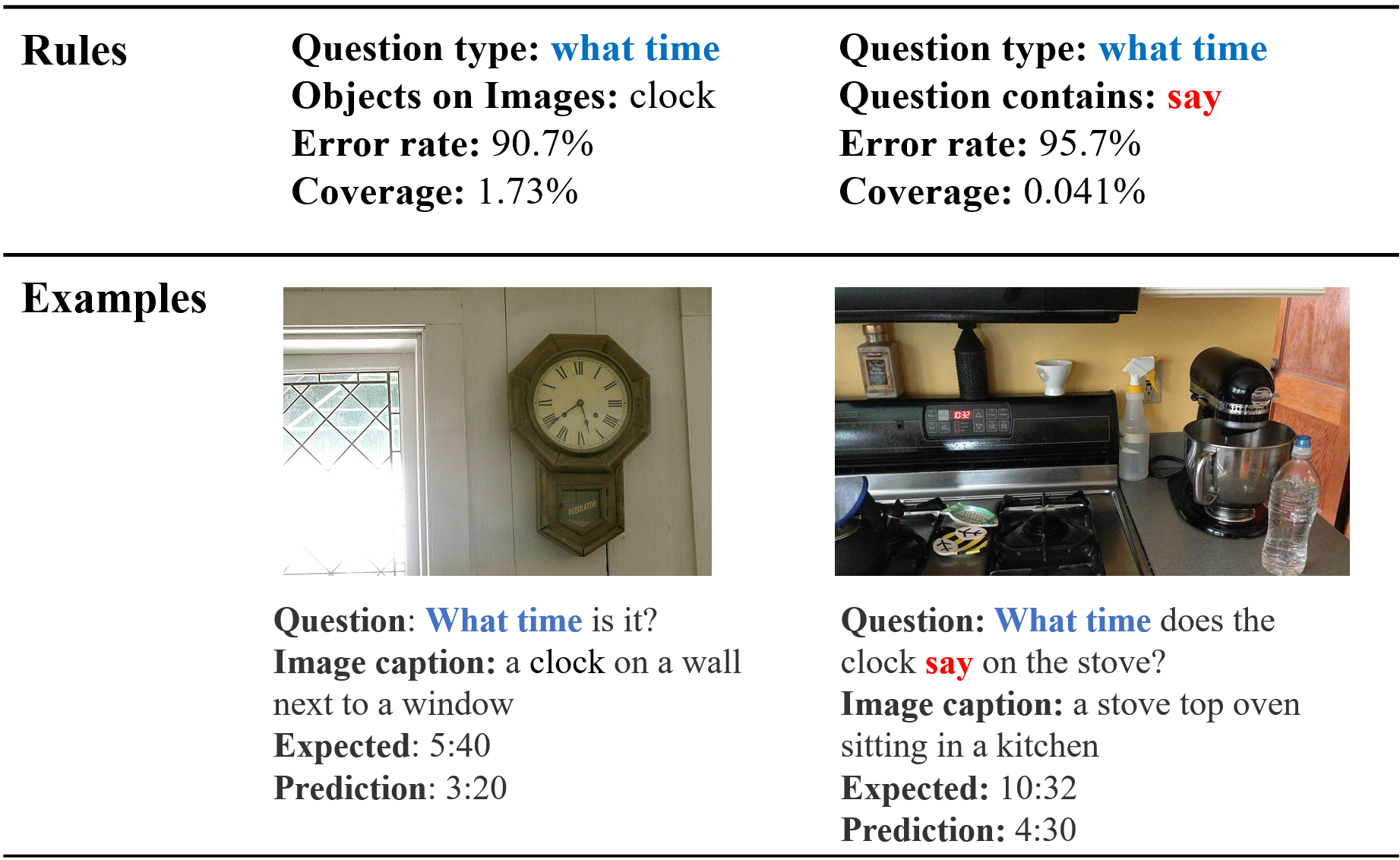}
    \caption{Time-reading problem examples}
    \label{fig:time}
\end{figure}
There are some rules found that cover ``what time'' questions. The rules in Figure \ref{fig:time} illustrates the challenge with these problems. Apart from the knowledge of various number representations, the underlying clock-reading rules are also difficult for the model to infer during training. 

\subsubsection{Counting Problems}
\begin{figure}[h]
    \centering
    \includegraphics[width=\columnwidth]{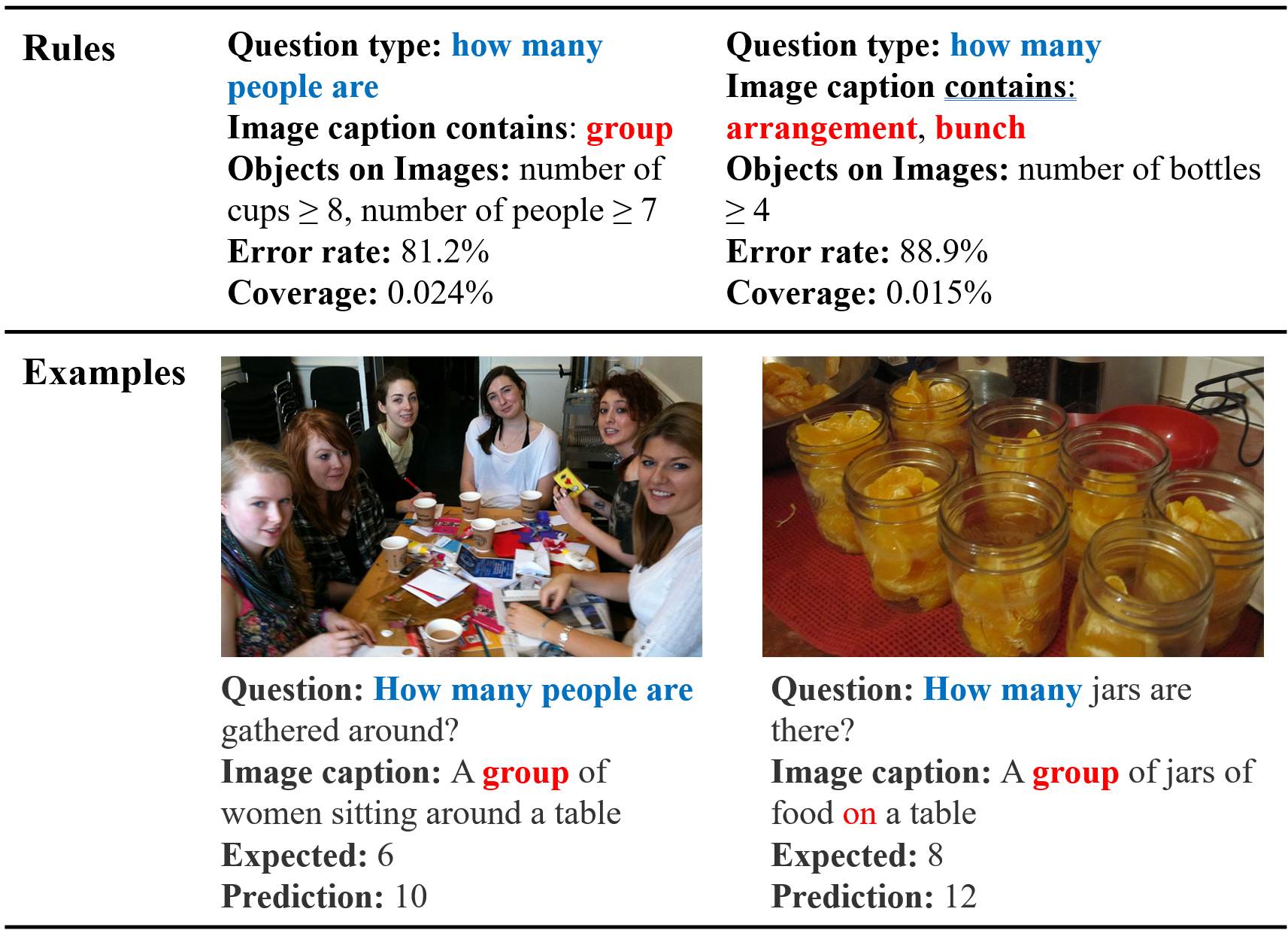}
    \caption{Counting problem examples}
    \label{fig:counting}
\end{figure}
VilBERT performs well on some counting problems; however, it fails on cluttered scenes with many objects as illustrated in Figure \ref{fig:counting}. 

\subsection{RoBERTa on Commonsense QA}
\begin{figure*}[h]
    \centering
    \includegraphics[width=0.8\textwidth]{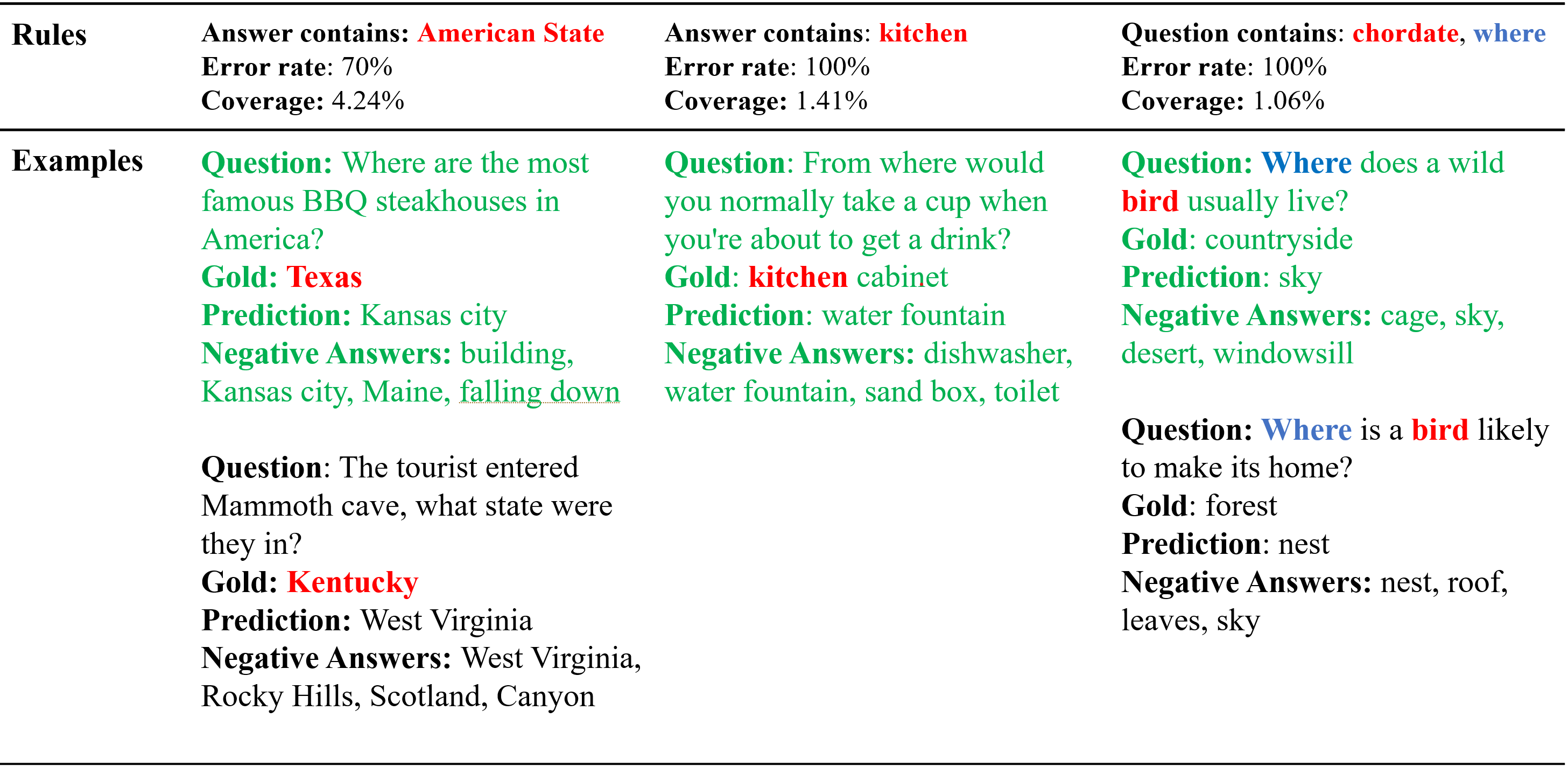}
    \caption{Rules for RoBERTa on CSQA, green examples are correctly answered by our ``refocused'' RoBERTa.}
    \label{fig:csqa}
\end{figure*}


We obtained several rules for RoBERTa on CSQA, that highlighted specific types of common sense knowledge which the model is lacking.
For instance, the rules learn that RoBERTa is unable to answer questions related to “American states”, suggesting that more prior knowledge about this topic would be helpful. More examples of concepts that seem to be the source of errors are shown in Figure \ref{fig:csqa}.

\subsection{Comparing Learning Approaches}
\label{sec:comp}

In order to evaluate the accuracy of the learned rules we use to predict error cases from meta-features, we compared our approach to other learned classifiers. We compared SkopeRules to decision trees, logistic regression, discriminative clustering \cite{Bansal2014TowardsTS} and a 2-layer MLP neural network.  
Decision trees, logistic regression and 2-layer MLP neural network follow the implementation in scikit-learn \citep{scikit-learn}. The maximum depth of the decision tree was set to 10. For the MLP network, the number of neurons of the first and the second hidden layers were 256 and 128 respectively.
We re-implemented the discriminative clustering method and chose the number of clusters to be 3.

\begin{figure}[h]
    \centering
    \includegraphics[width=0.8\columnwidth]{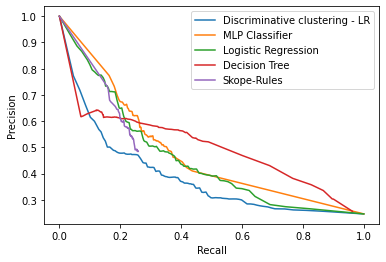}
    \caption{Precision-Recall curve of different classifiers}
    \label{fig:pr}
\end{figure}

Following the similar procedure in Section\ \ref{sec:ruleeval}, these classifiers were trained and evaluated on new training/test sets created from the original development set of VQA and CSQA. We averaged the classifiers' precision and recall weighted by size of datasets, and plotted an average precision-recall curve for predicting errors shown in Figure\ \ref{fig:pr}. 
The results show that our SkopeRules system is competitive with other models in terms of accurately predicting error cases, at least for the high precision end of the curve.\footnote{The curve of SkopeRules is truncated due to its high precision cut-off value in the training phase.} It is unable to achieve high recall; however, for aiding error analysis and interpreting the weaknesses of a model, we believe precisely modeling a subset of errors is more useful than achieving high recall with low precision. 

Although decision trees and discriminative clustering \citep{Bansal2014TowardsTS} are interpretable, the results show that they do not provide very high precision. While the weights on individual meta-features computed by logistic regression are also quite interpretable, we found that these globally computed feature importances without context do not provide as much insight. Conversely, SkopeRules considers combinations of meta-features under various cases, providing more interpretable insights than other competing methods.

Overall, we believe that learned rules provide the best trade-off between producing a high-precision model of error cases while also furnishing an interpretable characterization that provides the type of insight that one is typically seeking during error analysis.

\section{Improving Model Accuracy}

\label{sec:loop}
The ultimate goal of error analysis is use the insight gained about a model's weaknesses  to ``close the loop" and develop an improved model. Here we present our results on using the understanding gained from examining the rules learned by our approach to increase model accuracy.

\subsection{ViLBERT for VQA}

Not only do several learned rules indicate that ViLBERT has problems with reading text, we can use these rules to automatically identify cases where applying OCR might prevent the model from making a mistake. Therefore, we tried to utilize a 
 state-of-the-art OCR-VQA system, M4C \citep{hu2020iterative}, to improve ViLBERT.
 Specifically designed for OCR-VQA, M4C generates answers conditioning on additional text extracted by an OCR system. We used Mask TextSpotter v3 \citep{liao2020mask} as the OCR system. To ensure M4C fits the right distribution, we trained M4C on the official training set of VQA, then finetuned M4C on training examples whose images have text. During testing, we picked the 40 most accurate OCR rules (by F-score) to identify OCR questions, and redirect these questions to M4C. The final prediction of our system is a mixture of ViLBERT and M4C, depending on whether the rules indicate the question requires OCR. 


\begin{table*}[h]
\centering
\begin{tabular}{|l|l|llll|llll|}
\hline
Data Split    & \multicolumn{1}{c|}{dev} & \multicolumn{4}{c|}{test-dev}                                      & \multicolumn{4}{c|}{test-standard}                                 \\
\hline
Question Type & Overall & Yes/No         & Number         & Other          & Overall        & Yes/No         & Number         & Other          & Overall        \\
\hline
M4C           & 53.17 & 68.71          & 38.12          & 47.09          & 54.98          & 68.81          & 37.81          & 47.49          & 55.28          \\
ViLBERT       & 68.75 & \textbf{86.12} & 51.69          & 59.28          & 69.47          & \textbf{86.48} & 50.7           & 59.32          & 69.64          \\
Mix           & \textbf{69.44} & \textbf{86.12} & \textbf{52.84} & \textbf{59.35} & \textbf{69.64} & \textbf{86.48} & \textbf{51.79} & \textbf{59.46} & \textbf{69.82} \\
\hline
\end{tabular}
\caption{Performance on VQA v2.0}
\label{tab:accents}
\end{table*}
 
The results are in Table \ref{tab:accents} and show this ensemble is better than either of the individual systems. The improvement is modest due to the difficulty of OCR problems and relatively low global coverage of them in the dataset (only $3.86\%$ of the testing questions are identified as OCR problems).
However it illustrates how the learned rules can help direct system improvement.

\subsection{RoBERTa for CSQA}

The error-analysis rules learned for RoBERTa on CSQA point to a number of general conceptual areas where the model exhibits weaknesses.  One approach to improving the model based on this insight is to train the model on additional relevant data to correct for these deficiencies.
We observed that while Wikipedia is a part of the pretraining dataset, RoBERTa can still “forget” critical knowledge during large scale pretraining given the overwhelming coverage of Wikipedia. Hence, we introduce a dataset refinement approach to ensure that RoBERTa has used the given pre-training data efficiently and effectively for optimizing CSQA performance. 
We ranked discovered rules by their F-scores and picked 11 most interpretable rules for demonstration, which themselves consist of a keyword list as a summary of RoBERTa's ``knowledge gap.'' Next, we filtered Wikipedia data by retaining sentences containing these keywords or one of their hyponyms, as well as their neighboring sentences. Afterward, RoBERTa was first finetuned on this filtered Wikipedia data and then finetuned again on Commonsense QA, obtaining $78.7\%$ accuracy on the development set. 

As shown on Table \ref{tab:roberta}, our ``refocused'' RoBERTa obtained a 0.35 percentage point improvement on the CSQA test set, which is nearly the best among RoBERTa’s single model variants. We also show the best performance that the single RoBERTa model could achieve, which is trained with G-DAUG-Combo data augmentation technique \citep{yang-etal-2020-generative}\footnote{The actual best performing RoBERTa on the leaderboard is finetuned on Open Mind Common Sense (OMCS) corpus and has an accuracy of 73.3\%. However, since Commonsense QA is generated from ConceptNet, which is a part of OMCS, finetuning RoBERTa on OMCS may lead to unexpected data exposure and unfair comparison.}.
Compared to this method, we did not augment the dataset with complex synthetic data but only refocused training on critical aspects of the original training data and still obtained 70\% of the additional improvement gained by the best-performing model. The refocused model
corrects several of the original errors shown in Figure\ \ref{fig:csqa}, illustrating the effectiveness of our approach. 

\begin{table}[h]
\centering
    \begin{tabular}{|l|l|}
    \hline
    \textbf{Model} & \textbf{Accuracy} \\
    \hline
    RoBERTa + G-DAUG-Combo & 72.6 \\
    Ours & 72.45 \\
    RoBERTa (Baseline) & 72.1 \\
    \hline
    \end{tabular}
\caption{Single RoBERTa's performance on CSQA test split, collected from the public leaderboard compared to our ``re-focused'' model.}
\label{tab:roberta}
\end{table}

\section{Related Work}

\subsection{Black-box Explanation}
In the Explainable Artificial Intelligence (XAI) literature, the decisions made by a black-box model could be explained example-wise, i.e. in local explanations \citep{lime, shap}. Another way to explain a black-box model is disclosing its underlying decision logic in a structural and organized way, which form global explanations \citep{decisiontree, ribeiro2018anchors, palm}. Among these techniques, there has been a body of work on “rule extraction” from complex machine learning models that uses rule
induction methods to produce a surrogate model that can predict the output of the trained black-box model \citep{Shavlik1999RuleEW, sushilBlackBox18, manjunathaCVPR19, ramon2020metafeatures}. Unlike these prior works that extract rules mimicking the I/O behavior of a model, our approach
suggest a new way to understand black-box models, which attempts to predict when their output is wrong and directly support error analysis. This provides a different form of
“global explanation” that specifically characterizes the limitations of a model rather than just trying to predict its exact
behavior.

\subsection{Bias and Error Analysis}

As Machine Learning techniques have gained traction in the society, revealing their potential biases and errors are also becoming vitally important. Some works have already investigated the weaknesses of VQA models \citep{kafle2017analysis, agrawal-etal-2016-analyzing}, but the methods are highly specific to the task and irreproducible, whereas our method is task-agnostic given generic meta-features.
\citet{vqabiases} used a rule learner to imitate existing VQA models and explore their biases. However, rule learning is not an ideal approximator to the complex model, resulting in very limited insights. 
Similar to our work, \citet{rondeau-hazen-2018-systematic} designed meta-features associated with hypothesized challenges and trained a logistic regression classifier that predicts the difficulty of questions to evaluate hypotheses systematically. Errudite \citep{wu-etal-2019-errudite} helps users explore and test hypotheses to identify the underlying cause of failure. Both of them, however, require strong prior hypotheses about errors before users can design tests and gain in-depth insights from the results. In contrast, our approach only needs weak hypotheses on potentially influential properties in the meta-feature extraction process, which would be systematically and unbiasedly validated in batches, directing users to true patterns of errors and significantly facilitating the error analysis process. 
\citet{Bansal2014TowardsTS} use a discriminative clustering approach to characterize classes of errors made by an image classifier. Unlike the declarative rules learned by our approach, the discovered clusters can be difficult to interpret. We also extract a broader range of interpretable meta-features to characterize instances beyond the discretized image features employed in this approach, allowing a richer characterization of error classes. Moreover, none of aforementioned works has shown the practical value of error analysis,
while we build better systems guided by interpretable insights and proved the quality of them.

\subsection{Meta Features}
The term ``meta-feature'' in meta-learning is used to refer to features that characterize dataset complexity \cite{dataset-meta-1, jomaa2021dataset2vec, xia-etal-2020-predicting}. On the other hand, concurrent works in XAI use a different interpretation of the term. ``Super-pixels" in image classification explanation are sometimes used as  meta-features \citep{wei2018explain}. 
In the context of natural language classification, meta-features are created by either data-driven \citep{ramon2020metafeatures} or domain-based approaches \citep{SHEN, CHEN2016173} addressing the sparsity of textual features. Our approach utilizes domain-based meta-features, which categorize features into high-level and semantically similar groups even across modalities while maintaining their interpretability. These meta-features have broadened the coverage of rules and facilitated the error analysis process.


\section{Conclusion \& Future Work}

In this paper, we presented a novel pipeline that helps automate the error analysis process by learning interpretable rules that characterize the type of mistakes that a system makes. We then demonstrated its effectiveness by applying this pipeline to two different tasks. 
Presented in the form of well-organized rules, deficiencies and weaknesses in the model, datasets, and even the evaluation metrics were disclosed to researchers, shedding light on potential directions for improvement. We demonstrated the ability to ``close the loop'' and use the insight gained from some of the induced rules to make modest improvements to these systems. These simple but effective approaches have the potential to be applied in production environment shortening the iterative update cycle of models.

As discussed in Section\ \ref{sec:exp}, our analysis revealed several other problems for these tasks that remain open and could be addressed in future development. 
Besides, there are a number of aspects also deserving of future study before the achievement of full automated error analysis. For example, although the requirement for manual customization to tasks is already quite limited in our error analysis approach, meta-feature extraction is still a critical step that is responsible for disclosing underlying interrelationships among error examples and determining their human understandability.
Therefore, apart from using hand-crafted meta-features only, data-driven meta-features could also be explored, though it remains challenging how to ensure that data-driven meta-features are still understandable by human developers. 

In addition, incorporating such automated error analysis into an overall automated machine learning (AutoML) process is another promising direction for future work. Most of the AutoML literature \citep{bay-automl, white2020local, nat} heavily relies on a single metric to evaluate the model's performance, whereas such an individual metric is usually uninformative of errors and biases in the model, limiting the development process. Our pipeline enriches the information available about weaknesses of the current model and could potentially aid a process that helps automatically ``close the loop'' and improve the speed and effectiveness of the evolution of continually improved models.

\section{Acknowledgment}
This research was supported by the DARPA XAI program under a grant from AFRL. We would like to thank Bill Ferguson and his colleagues at Raytheon for inspiring our work on global explanations.

\bibliography{aaai22}
\newpage
\section{Appendix}
\label{sec:appendix}
\subsection{Other Error Characterizations Detected on VQA}
\subsubsection{Commonsense Knowledge}
\begin{figure*}[h]
    \centering
    \includegraphics[width=0.8\textwidth]{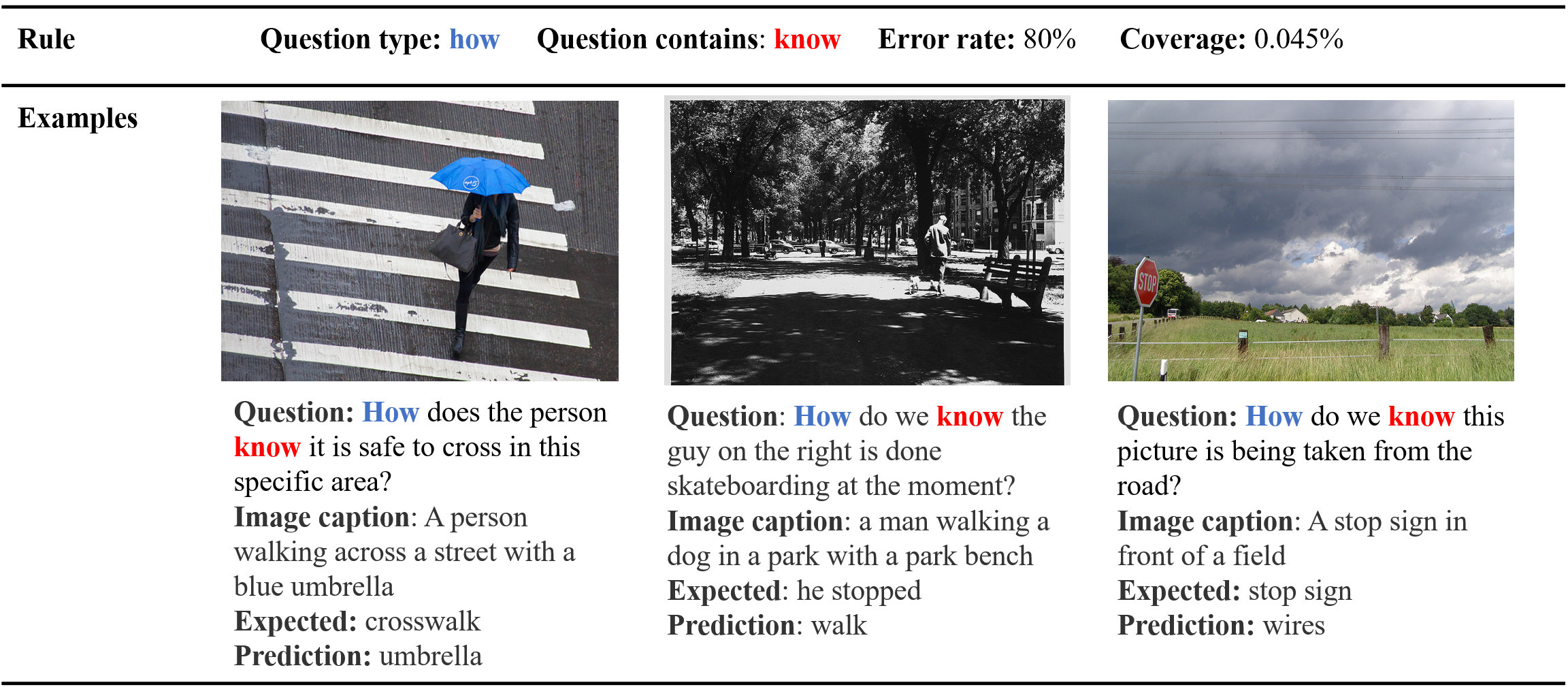}
    \caption{Commonsense problem examples}
    \label{fig:vqa_cs}
\end{figure*}

“How” questions with “know” word are likely commonsense questions that the model could not answer. 
While ViLBERT often answers such questions with nonsensical words, we observed some cases that reasonable answers could not get points in evaluation, as shown in the second example in Figure \ref{fig:vqa_cs}.
Similar situation also happens in ``why"-type questions, where the content of questions and answers are totally unrestricted and impose challenges on both the model and the current evaluation metric.

\begin{figure*}[h]
    \centering
    \includegraphics[width=0.8\textwidth]{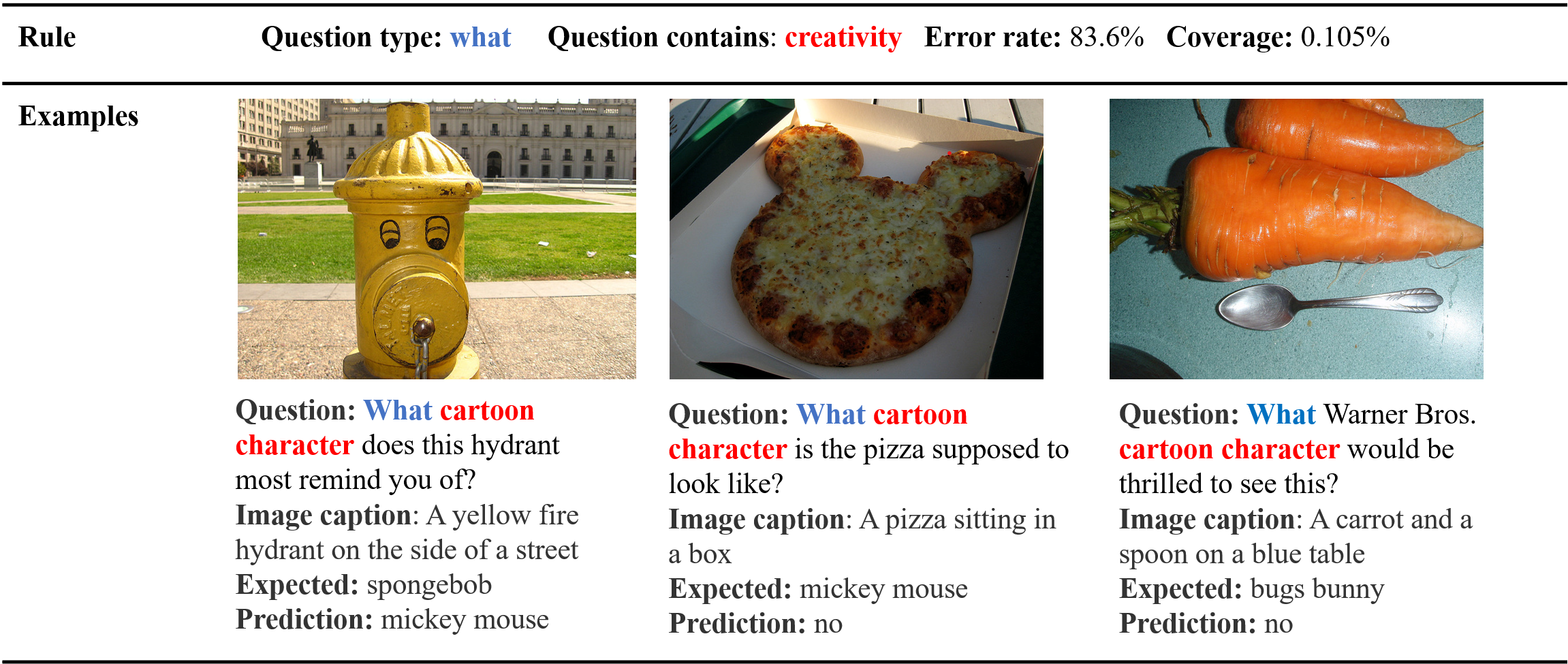}
    \caption{Ficitional problem examples}
    \label{fig:vqa_ficitional}
\end{figure*}
\subsubsection{Fictional Characters}
ViLBERT fails on “what” questions with words whose hypernym is “creativity”. These questions ask the model to identify or reason the cartoon characters on the image. The first two examples in Figure\ \ref{fig:vqa_ficitional} indicate that difficulty lies in learning the mapping between abstract 2D scratches to natural 3D objects as their details usually vary significantly, and ViLBERT did not demonstrate such ability. The third example even needs the model to speculate the future acting as a character, whereas no hint could be found in the training set, suggesting that insufficient training data on certain domain is another critical factor to ViLBERT's failure.

\begin{figure*}[h]
    \centering
    \includegraphics[width=0.8\textwidth]{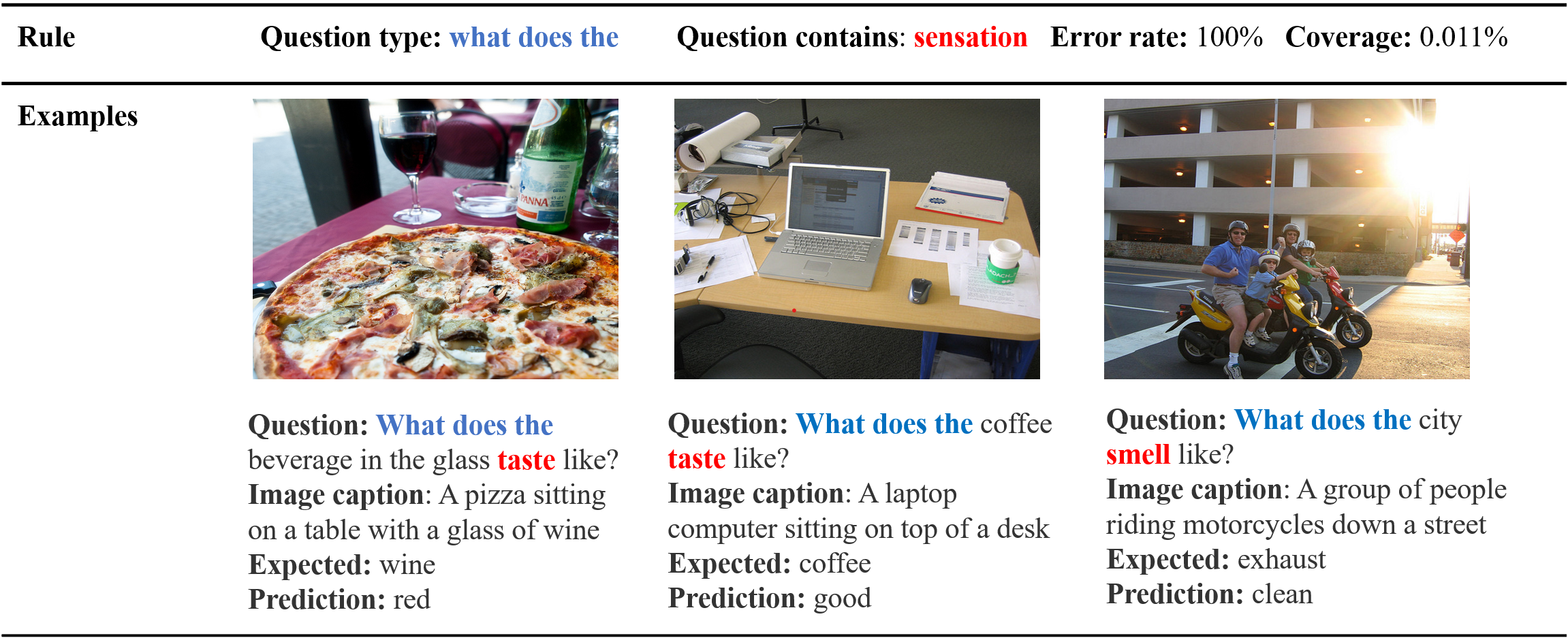}
    \caption{Sensation problem examples}
    \label{fig:vqa_sensation}
\end{figure*}
\subsubsection{Sensation Problems}
One interesting rule shown in Figure \ref{fig:vqa_sensation} detects an odd set of problems involving human sensory judgement (e.g. taste, smell).
The gold answers for these problems are also dubious or cover only a subset of reasonable answers.  We believe most of these problems are just poor choices for VQA, indicating a problem with the data rather than the model.    
\subsection{Rules Used for Improvement}
Here we present raw rules used to guide the models' improvement in the main paper for CSQA (table \ref{table:csqa_rules}) and VQA (table \ref{table:vqa_rules}).

\begin{table}[h]
\centering
\begin{tabular}{|l|}
\hline
\textbf{Rule}                                                                                  \\\hline
answer\_american\_state $>$ 0.5                                              \\\hline
question\_educational\_institution $>$ 0.5 and \\
question\_to $>$ 0.5 \\\hline
answer\_natural\_phenomenon $>$ 0.5                                          \\\hline
answer\_mishap $>$ 0.5                                                       \\\hline
answer\_kitchen $>$ 0.5                                                      \\\hline
answer\_cupboard $>$ 0.5                                                     \\\hline
question\_rational\_motive $>$ 0.5                                           \\\hline
answer\_geographical\_area $>$ 0.5 and \\
question\_location $>$ 0.5   \\\hline
question\_television $>$ 0.5                                                 \\\hline
answer\_juvenile $>$ 0.5                                                     \\\hline
question\_astonishment $>$ 0.5                                              
\\\hline
\end{tabular}
\caption{Rule picked for RoBERTa. Words in these rules are directly used to filter the Wikipedia dataset.}
\label{table:csqa_rules}
\end{table}

\begin{table*}[h]
\centering
\begin{tabular}{|l|l|}
\hline
\textbf{Question Type} & \textbf{Rule} \\ \hline
what & image\_caption\_evidence $>$ 0.5\\ \hline 
what does the & question\_say $>$ 0.5\\ \hline 
what is the & question\_magnitude $>$ 0.5\\ \hline 
what does the & question\_evidence $>$ 0.5\\ \hline 
what & question\_social\_group $>$ 0.5\\ \hline 
what & question\_say $>$ 0.5\\ \hline 
what number is & image\_object\_person $>$ 2.5\\ \hline
what number is & image\_caption\_matter $>$ 0.5\\ \hline
what & question\_document $>$ 0.5\\ \hline
what is the name & image\_caption\_way $>$ 0.5\\ \hline
what & question\_language\_unit $>$ 0.5\\ \hline
what number is & question\_instrumentality $>$ 0.5\\ \hline
what is the name & image\_caption\_evidence $>$ 0.5\\ \hline 
what number is & image\_object\_person $>$ 5.5\\ \hline
what number is & question\_contestant $>$ 0.5\\ \hline
what are the & question\_number $>$ 0.5\\ \hline
what number is & image\_caption\_state $>$ 0.5\\ \hline
what number is & image\_caption\_facility $>$ 0.5\\ \hline
what number is & image\_caption\_facility $>$ 0.5\\ \hline 
what is the & question\_motor\_vehicle $>$ 0.5\\ \hline 
what number is & image\_caption\_two $>$ 0.5\\ \hline 
what is the & question\_stopword $>$ 0.5 and question\_use $>$ 1.0\\ \hline
what number is & image\_caption\_consumer\_goods $>$ 0.5\\ \hline
what number is & image\_caption\_measure $>$ 0.5\\ \hline
what does the & image\_caption\_back $>$ 0.5\\ \hline
what & question\_act $>$ 0.5 and question\_year $>$ 1.0\\ \hline
which & question\_relation $>$ 0.5\\ \hline
what are the & question\_letter $>$ 1.0\\ \hline
what number is & image\_caption\_rid $>$ 0.5\\ \hline
what kind of & question\_product $>$ 0.5\\ \hline
what is the & image\_object\_clock $>$ 1.5 and question\_time $>$ 1.0\\ \hline
what is the & question\_company $>$ 1.0\\ \hline
what does the & image\_caption\_proctor $>$ 0.5\\ \hline
where are the & image\_caption\_radiation $>$ 0.5\\ \hline
what is & question\_truck $>$ 1.0\\ \hline
what is & image\_caption\_travel $>$ 0.5 and question\_announce $>$ 0.5\\ \hline
what is the & image\_caption\_vascular\_plant $>$ 0.5 and question\_bus $>$ 1.0\\ \hline
what is in the & image\_object\_hot\_dog $>$ 0.5\\ \hline
what is in the & image\_caption\_unpleasant\_person $>$ 0.5\\ \hline
what is this & image\_object\_boat $>$ 1.5\\ \hline
\end{tabular}
\caption{OCR Rules used for ViLBERT's improvement}
\label{table:vqa_rules}
\end{table*}


\end{document}